\documentclass[journal]{IEEEtran}

\usepackage{cite}
\usepackage{url}
\usepackage{graphicx}
\usepackage[caption=false,font=footnotesize]{subfig}
\usepackage{amsmath,amssymb}
\usepackage{bm}
\usepackage{booktabs}
\usepackage{array}
\usepackage{xcolor}
\usepackage{textcomp}
\usepackage{epstopdf}
\usepackage{balance}
\usepackage{microtype}
\usepackage{dblfloatfix}
\usepackage[ruled,vlined]{algorithm2e}

\usepackage{tikz}
\usetikzlibrary{matrix,fit,positioning,calc}
\newcommand{\twodots}{\vbox{\baselineskip=2.5pt \lineskiplimit=0pt \hbox{.}\hbox{.}}}
\SetAlgoCaptionSeparator{:}
\SetAlgoVlined
\DontPrintSemicolon
\SetKwInput{KwIn}{Input}
\SetKwInput{KwOut}{Output}
\SetKw{KwRet}{Return}
\SetAlFnt{\footnotesize}
\SetAlCapFnt{\footnotesize}
\SetAlCapNameFnt{\footnotesize\bfseries}
\SetInd{0.45em}{0.9em}

\AtBeginDocument{%
	\setlength{\textfloatsep}{7pt plus 1pt minus 2pt}%
	\setlength{\floatsep}{6pt plus 1pt minus 2pt}%
	\setlength{\intextsep}{6pt plus 1pt minus 2pt}%
	\setlength{\dbltextfloatsep}{7pt plus 1pt minus 2pt}%
	\setlength{\dblfloatsep}{6pt plus 1pt minus 2pt}%
}

\begin{document}
	
	\title{ Variational Outlier-Robust Gaussian Process Regression with Generative  Modeling}
	
	\author{Arslan~Majal and Aamir~Hussain~Chughtai%
		\thanks{A. Majal is with Zunitech, LLC, Houston, USA.}
		\thanks{A. H. Chughtai is with the Institute of Data Science, University of Engineering and  Technology, Lahore, Pakistan.}%
	}
	
	\markboth{IEEE Signal Processing Letters,~Vol.~XX, No.~XX, 2026}%
	{Majal \MakeLowercase{\textit{et al.}}: ASOR-GPR for Outlier-Robust Gaussian Process Regression}
	
	\maketitle

	\begin{abstract}
		Outliers can substantially distort Gaussian process regression (GPR)
		due to its conventional Gaussian observation likelihood, leading to
		inaccurate model learning and prediction. To address this limitation,
		this article introduces a generative GPR model that captures
		observation-specific contamination and adaptively mitigates the
		influence of outliers. Subsequently, a variational generalized
		expectation-maximization procedure is used to learn the latent
		variables and GPR model parameters. Experiments on synthetic and real
		datasets under different contamination settings demonstrate that the
		proposed method remains competitive with---and in several cases
		outperforms---robust GPR baselines in prediction accuracy. Moreover, the proposed method shares the cubic computational scaling
of the compared GPR methods.
	\end{abstract}

	\begin{IEEEkeywords}
		Gaussian process regression, outlier-robust regression, variational Bayesian inference, robust signal processing, uncertainty quantification, expectation-maximization
	\end{IEEEkeywords}

	\IEEEpeerreviewmaketitle

	\section{Introduction}
	
	\IEEEPARstart{G}{aussian} process regression (GPR)
	\cite{rasmussen2006gaussian} is a Bayesian nonparametric regression
	method that provides flexible nonlinear modeling and principled
	uncertainty quantification, while remaining effective in data-limited
	settings. These properties make GPR useful in numerous
	signal-processing and learning applications.
	
	Standard GPR assumes Gaussian observation noise, making it sensitive
	to outliers caused by sensor faults, calibration errors, transmission
	noise, or unexpected disturbances. Under this model, large residuals
	are penalized quadratically, allowing even a few corrupted
	observations to substantially distort the estimated latent function
	\cite{huber1964robust}. This motivates robust GPR methods that reduce
	the influence of unreliable observations.
	
	Outliers in GPR are commonly handled by modifying the observation
	likelihood or the corresponding loss function. Existing robust
	approaches often rely on fixed likelihood modifications or robust
	losses with user-defined tuning parameters, including Huber- and
	Hampel-type penalties, heavy-tailed Student-$t$ likelihoods, and
	generalized-Bayes likelihood-induced losses
	\cite{7004783,jylanki2011robust,altamirano2024robust}.
	Although tractable, these approaches can be sensitive to prescribed
	thresholds, loss parameters, or distributional shape parameters that
	control the influence of atypical observations. To overcome these limitations, more expressive adaptive models have
	been proposed. Instead of fixing the robust cost or likelihood
	parameters a priori, these methods adapt them from the observed data,
	thereby providing data-driven robustness.
	
	The main challenge,
	however, is to make the outlier model sufficiently expressive while
	retaining tractable inference. To this end, existing approaches
	introduce latent variables, hierarchical likelihoods, mixture-based
	noise models, or optimization-based robust formulations, including
	graduated non-convexity (GNC), to account for corrupted
	measurements
	\cite{wang2018robust,9716131,nakabayashi2019nonlinear,
		daemi2019gaussian,10430181,10679915,11204236,Yang20ral-GNC}.
	
	A relevant class of adaptive methods employs generative
	models with variational inference to identify and mitigate outliers \cite{9716131,10679915}. Building on this idea, adaptive
	selective observation rejecting (ASOR) \cite{10430181} learns outlier
	statistics rather than completely discarding suspected observations,
	thereby adapting their influence according to the learned
	contamination characteristics. ASOR has demonstrated favorable estimation accuracy and computational
	efficiency relative to state-of-the-art GNC methods in robotic
	perception problems \cite{Yang20ral-GNC}. Its conjugate hierarchical
	structure enables closed-form variational updates for the model parameters \cite{10430181}, with
	related ideas extended to robust filtering in \cite{11204236}.

	Motivated by the adaptive outlier treatment in ASOR, we propose a formulation tailored to outlier-robust GPR. The key
	development is to integrate observation-specific precision modeling
	with the latent GP function and its associated noise, mean, and kernel
	parameters within a unified model. Since exact Bayesian
	inference is intractable, we employ a variational
	generalized expectation-maximization (EM) procedure to learn the latent
	variables and model parameters. Closed-form updates are obtained for all quantities except the kernel
	hyperparameters, which are optimized using the gradient-based method
	\cite{daemi2019gaussian}.  
	
	To evaluate performance, the proposed method is extensively compared
	with representative GPR baselines on synthetic and real-data
	regression problems under diverse contamination settings.	
	\section{Proposed Methodology}
	\label{sec:proposed_methodology}
	\emph{Notation:}
	Nonbold symbols denote scalars or scalar-valued functions, bold
	lowercase symbols denote vectors, and bold uppercase symbols denote
	matrices. Moreover, $(\cdot)^{\top}$ denotes transpose,
	$[\boldsymbol{A}]_{ij}$ denotes the $(i,j)$th entry of
	$\boldsymbol{A}$, and $\operatorname{diag}(\cdot)$ and
	$\operatorname{tr}(\cdot)$ denote the diagonal and trace operators,
	respectively.
	
	Consider $n$ input--output samples arranged as
	\begin{equation*}
		\resizebox{0.88\columnwidth}{!}{%
			\begin{tikzpicture}[
				baseline=(Xmat.center),
				font=\scriptsize,
				xmat/.style={
					matrix of math nodes,
					ampersand replacement=\&,
					left delimiter={[},
					right delimiter={]},
					nodes={
						minimum width=0.56cm,
						minimum height=0.31cm,
						anchor=center
					},
					column sep=0pt,
					row sep=0pt
				},
				ymat/.style={
					matrix of math nodes,
					ampersand replacement=\&,
					left delimiter={[},
					right delimiter={]},
					nodes={
						minimum width=0.40cm,
						minimum height=0.31cm,
						anchor=center
					},
					column sep=0pt,
					row sep=0pt
				}
				]
				
				\matrix (Xmat) [xmat] {
					x_{11} \& \cdots \& x_{1d_x} \\
					\twodots \&        \& \twodots \\
					x_{i1} \& \cdots \& x_{id_x} \\
					\twodots  \&        \& \twodots \\
					x_{n1} \& \cdots \& x_{nd_x} \\
				};
				
				\matrix (Ymat) [
				ymat,
				right=12mm of Xmat.east
				] {
					y_{11} \& \cdots \& y_{1j} \& \cdots \& y_{1d_y} \\
					\twodots \&        \& \twodots \&        \& \twodots  \\
					y_{i1} \& \cdots \& y_{ij} \& \cdots \& y_{id_y} \\
					\twodots  \&        \& \twodots  \&        \& \twodots \\
					y_{n1} \& \cdots \& y_{nj} \& \cdots \& y_{nd_y} \\
				};
				
				% Matrix labels
				\node[
				anchor=east,
				inner sep=0pt
				] at ([xshift=-2.8mm,yshift=-6.2mm]Xmat.north west)
				{$\boldsymbol{X}=$};
				
				\node[
				anchor=east,
				inner sep=0pt
				] at ([xshift=-1.8mm,yshift=-6.9mm]Ymat.north west)
				{$\boldsymbol{Y}=$};
				
				\node[ inner sep=0pt ] at ($(Xmat.east)!0.3!(Ymat.west)$) {$,$};

				% Highlighted ith row of X
				\node[
				draw=red,
				dashed,
				rounded corners=1.2pt,
				inner sep=0.8pt,
				fit=(Xmat-3-1)(Xmat-3-3)
				] {};
				
				\node[
				text=red,
				anchor=east,
				inner sep=0pt
				] at ([xshift=-1.2mm]Xmat.west |- Xmat-3-1.center)
				{$(\boldsymbol{x}^{i})^{\top}$};
				
				% Highlighted ith row of Y
				\node[
				draw=red,
				dashed,
				rounded corners=1.2pt,
				inner sep=0.8pt,
				fit=(Ymat-3-1)(Ymat-3-5)
				] {};
				
				\node[
				text=red,
				anchor=west,
				inner sep=0pt
				] at ([xshift=1.2mm]Ymat.east |- Ymat-3-5.center)
				{$(\boldsymbol{y}^{i})^{\top}$};
				
				% Highlighted jth column of Y
				\node[
				draw=blue,
				densely dotted,
				rounded corners=1.2pt,
				inner sep=0.8pt,
				fit=(Ymat-1-3)(Ymat-5-3)
				] {};
				
				\node[
				text=blue,
				anchor=north,
				inner sep=0pt
				] at ([yshift=-0.7mm]Ymat-5-3.south)
				{$\boldsymbol{y}_j$};
				
			\end{tikzpicture}%
		}
		\label{eq:data_matrix_views}
	\end{equation*}where $\boldsymbol{X}\in\mathbb{R}^{n\times d_x}$ contains the input
	samples and
	$\boldsymbol{Y}\in\mathbb{R}^{n\times d_y}$ contains the corresponding
	outputs. The $i$th rows are
	$(\boldsymbol{x}^{i})^{\top}$ and
	$(\boldsymbol{y}^{i})^{\top}$, respectively, while
	$\boldsymbol{y}_j$ denotes the $j$th output column.
	
	Assuming independent output components and contamination processes,
	the problem is decomposed into $d_y$ scalar-output GPR models, where
	the $j$th model is trained using $\boldsymbol{X}$ and
	$\boldsymbol{y}_j$.

	\subsection{Scalar Observation Model}
	
	For output dimension $j$, let
	$f_{ij}\triangleq f_j(\boldsymbol{x}^{i})$ denote the latent function
	value at the $i$th input. The observation model is
	\begin{equation*}
		y_{ij}
		=
		f_{ij}
		+
		\varepsilon_{ij},
		\quad
		i=1,\ldots,n,
	\end{equation*}
	where $\varepsilon_{ij}$ denotes the observation error and
	$\sigma_j^2$ denotes its nominal variance. Following the ASOR
	construction, a positive latent precision multiplier
	$\mathcal{I}_{ij}>0$ is introduced such that
	\[
	p(\varepsilon_{ij}\mid\mathcal{I}_{ij},\sigma_j^2)
	=
	\mathcal{N}
	\left(
	\varepsilon_{ij}
	\mid
	0,
	{\sigma_j^2}/{\mathcal{I}_{ij}}
	\right),
	\]
	where $\mathcal{N}(x\mid\mu,v)$ denotes a Gaussian density in $x$
	with mean $\mu$ and variance $v$. Accordingly, the observation
	likelihood is
	\begin{equation}
		p(y_{ij}\mid f_{ij},\mathcal{I}_{ij},\sigma_j^2)
		=
		(2\pi\sigma_j^2)^{-1/2}
		\mathcal{I}_{ij}^{1/2}
		\exp\!\left[
		-\frac{\mathcal{I}_{ij}}{2\sigma_j^2}
		(y_{ij}-f_{ij})^2
		\right].
		\label{eq:scalar_likelihood_explicit}
	\end{equation}
	The value $\mathcal{I}_{ij}=1$ recovers the nominal observation model,
	whereas $\mathcal{I}_{ij}\neq1$ adaptively modifies the observation
	variance to account for outliers.
	
	Let
	$\boldsymbol{f}_j\triangleq
	[f_{1j},\ldots,f_{nj}]^{\top}$. The GP prior is
	\begin{equation}
		\begin{aligned}
			p(\boldsymbol{f}_j
			\mid\boldsymbol{X},m_j,\boldsymbol{\kappa}_j)
			&=
			\mathcal{N}
			\left(
			\boldsymbol{f}_j
			\mid
			m_j\boldsymbol{1}_n,
			\boldsymbol{K}_j
			\right),
			\\
			[\boldsymbol{K}_j]_{i\ell}
			&=
			k_j(\boldsymbol{x}^{i},\boldsymbol{x}^{\ell};
			\boldsymbol{\kappa}_j).
		\end{aligned}
		\label{eq:gp_prior}
	\end{equation}
	where $\boldsymbol{1}_n\in\mathbb{R}^{n}$ denotes the vector of ones
	and, for the multivariate case,
	$\mathcal{N}(\boldsymbol{x}\mid\boldsymbol{\mu},\boldsymbol{\Sigma})$
	denotes a Gaussian density with mean $\boldsymbol{\mu}$ and covariance
	matrix $\boldsymbol{\Sigma}$. The covariance matrix is constructed using the automatic relevance
	determination (ARD) squared-exponential kernel
	\begin{equation}
		\begin{aligned}
			k_j(\boldsymbol{x}^{i},\boldsymbol{x}^{\ell};
			\boldsymbol{\kappa}_j)
			&=
			\sigma_{f,j}^{2}
			e^{
				-0.5
				\sum_{r=1}^{d_x}
				{(x_{ir}-x_{\ell r})^2}/{\ell_{jr}^{2}}
			},
			\\
			\boldsymbol{\kappa}_j
			&\triangleq
			[\sigma_{f,j}^{2},\ell_{j1},\ldots,
			\ell_{jd_x}]^{\top}.
		\end{aligned}
		\label{eq:ard_se_kernel}
	\end{equation}
	Here, $\sigma_{f,j}^{2}$ is the signal variance and $\ell_{jr}$ is
	the length scale associated with the $r$th input dimension.
	
	\subsection{Hierarchical Priors}
	
	$\mathcal{I}_{ij}$ is assigned the unit-spike-and-Gamma prior
	\begin{equation}
		p(\mathcal{I}_{ij}\mid b_j)
		=
		(1-\theta_{ij})
		\mathcal{G}(\mathcal{I}_{ij}\mid a,b_j)
		+
		\theta_{ij}
		\delta(\mathcal{I}_{ij}-1),
		\label{eq:latent_precision_prior}
	\end{equation}
	where $\delta(\cdot)$ denotes the Dirac delta function and
	$\mathcal{G}(\mathcal{I}\mid a,b)$ denotes the Gamma density with
	shape $a>0$ and rate $b>0$:
	\begin{equation}
		\mathcal{G}(\mathcal{I}\mid a,b)
		\triangleq
		\frac{b^a}{\Gamma(a)}
		\mathcal{I}^{a-1}
		e^{-b\mathcal{I}}.
		\label{eq:gamma_definitions}
	\end{equation}
	where $\Gamma(\cdot)$ denotes the Gamma function. The parameter $\theta_{ij}\in(0,1)$ is the prior probability that
	$y_{ij}$ belongs to the nominal branch. 
	
	To retain Bayesian conjugacy, the output-specific Gamma rate
	$b_j$ is assigned the Gamma hyperprior
	\begin{equation}
		p(b_j)
		=
		\mathcal{G}(b_j\mid A,B),
		\label{eq:b_prior}
	\end{equation}
	where $A>1$ and $B>0$ are fixed hyperparameters.
	
	Similarly, the nominal observation-noise variance is assigned the
	inverse-Gamma prior
	\begin{equation}
		\begin{aligned}
			p(\sigma_j^2)
			&=
			\mathcal{IG}
			\left(
			\sigma_j^2
			\middle|
			\frac{\nu_0}{2},
			\frac{s_{0j}}{2}
			\right)\propto
			(\sigma_j^2)^{-(\nu_0+2)/2}
			e^{-{s_{0j}}/{2\sigma_j^2}}.
		\end{aligned}
		\label{eq:sigma_prior}
	\end{equation}
	where $\nu_0>0$ and $s_{0j}>0$ are fixed hyperparameters.
	
	The fixed prior hyperparameters associated with output dimension
	$j$ are collected as
	\begin{equation}
		\mathcal{P}_j
		\triangleq
		\left\{
		a,
		\{\theta_{ij}\}_{i=1}^{n},
		A,
		B,
		\nu_0,
		s_{0j}
		\right\}.
		\label{eq:fixed_prior_hyperparameters}
	\end{equation}
	
	\subsection{Variational Generalized-EM Inference}
	
	For output dimension $j$, define
	\begin{equation}
		\boldsymbol{\mathcal{I}}_j
		\triangleq
		[\mathcal{I}_{1j},\ldots,\mathcal{I}_{nj}]^{\top},
		\;
		\boldsymbol{Z}_j
		\triangleq
		\{\boldsymbol{f}_j,\boldsymbol{\mathcal{I}}_j\},
		\;
		\boldsymbol{\Theta}_j
		\triangleq
		\{m_j,\boldsymbol{\kappa}_j,\sigma_j^2,b_j\}.
		\label{eq:latent_and_parameter_sets}
	\end{equation}
	
	For $j=1,\ldots,d_y$, the joint posterior satisfies
	\begin{equation}
		\begin{aligned}
			p(\boldsymbol{Z}_j,\boldsymbol{\Theta}_j
			\mid\boldsymbol{y}_j,\boldsymbol{X},\mathcal{P}_j)
			&\propto
			p(\boldsymbol{f}_j
			\mid\boldsymbol{X},m_j,\boldsymbol{\kappa}_j)
			p(\sigma_j^2)p(b_j)
			\\
			&\times
			\prod_{i=1}^{n}
			p(\mathcal{I}_{ij}\mid b_j)
			p(y_{ij}\mid f_{ij},\mathcal{I}_{ij},\sigma_j^2).
		\end{aligned}
		\label{eq:joint_posterior_all_unknowns}
	\end{equation}
    
	We use flat non-informative priors for $m_j$ and
	$\boldsymbol{\kappa}_j$, whose constant terms are omitted from
	\eqref{eq:joint_posterior_all_unknowns}. Since the exact joint posterior does not admit closed form, we employ variational generalized EM
	\cite{vsmidl2006variational}. 
	
	The E-step uses the mean-field
	factorization
	\begin{equation}
		q_j(\boldsymbol{Z}_j)
		=
		q_j(\boldsymbol{f}_j)
		\prod_{i=1}^{n}q_j(\mathcal{I}_{ij}).
		\label{eq:latent_variational_factorization}
	\end{equation}
    
	For
	$\xi\in\{\boldsymbol{f}_j,\mathcal{I}_{1j},...,
	\mathcal{I}_{nj}\}$, the coordinate-optimal update is
	\begin{equation}
		\log q_j^{\star}(\xi)
		=
		\mathbb{E}_{q_j(\boldsymbol{Z}_j\setminus\xi)}
		\!\left[
		\log p\!\left(
		\boldsymbol{y}_j,\boldsymbol{Z}_j
		\mid
		\boldsymbol{X},\boldsymbol{\Theta}_j,\mathcal{P}_j
		\right)
		\right]
		+c,
		\label{eq:variational_coordinate_update}
	\end{equation}
	where $c$ is independent of $\xi$. Applying
	\eqref{eq:variational_coordinate_update} to
	$\boldsymbol{f}_j$ and each $\mathcal{I}_{ij}$ gives the closed-form
	updates below. At each generalized-EM iteration, these factors are updated
	sequentially, and their product gives $q_j(\boldsymbol{Z}_j)$.
	
	The M-step aims to solve
	\begin{equation}
		\boldsymbol{\Theta}_j^{\star}
		\in
		\underset{\boldsymbol{\Theta}_j}{\arg\max}\;
		\mathcal{Q}_j(\boldsymbol{\Theta}_j).
		\label{eq:M_step_argmax}
	\end{equation}
	where $\mathcal{Q}_j(\boldsymbol{\Theta}_j)
	\triangleq
	\mathbb{E}_{q_j(\boldsymbol{Z}_j)}
	\!\left[
	\log p\!\left(
	\boldsymbol{y}_j,
	\boldsymbol{Z}_j,
	\boldsymbol{\Theta}_j
	\mid
	\boldsymbol{X},
	\mathcal{P}_j
	\right)
	\right]$. Closed-form updates are obtained for $m_j$, $\sigma_j^2$, and $b_j$,
	whereas $\boldsymbol{\kappa}_j$ is updated using gradient
	descent with backtracking. The priors on $\sigma_j^2$ and $b_j$ yield  maximum a posteriori (MAP)
	updates, whereas the uninformative priors on $m_j$ and
	$\boldsymbol{\kappa}_j$ yield maximum-likelihood updates.
	
	\subsection{Variational E-Step: Latent-Function Update}
	
	For fixed output $j$, define
	\begin{equation}
		w_{ij}
		\triangleq
		\mathbb{E}_{q_j(\mathcal{I}_{ij})}
		[\mathcal{I}_{ij}],
		\;
		\boldsymbol{\Lambda}_j
		\triangleq
		\operatorname{diag}(w_{1j},\ldots,w_{nj})/{\sigma_j^2}.
		\label{eq:asor_weight_and_precision}
	\end{equation}
	Applying \eqref{eq:variational_coordinate_update} to
	$\boldsymbol{f}_j$ gives
	\begin{equation}
		\begin{aligned}
			q_j(\boldsymbol{f}_j)
			&=
			\mathcal{N}
			\left(
			\boldsymbol{f}_j
			\mid
			\boldsymbol{\mu}_{f,j},
			\boldsymbol{C}_{f,j}
			\right),
			\\
			\boldsymbol{C}_{f,j}
			&=
			\left(
			\boldsymbol{K}_j^{-1}
			+
			\boldsymbol{\Lambda}_j
			\right)^{-1},
			\\
			\boldsymbol{\mu}_{f,j}
			&=
			\boldsymbol{C}_{f,j}
			\left(
			\boldsymbol{K}_j^{-1}m_j\boldsymbol{1}_n
			+
			\boldsymbol{\Lambda}_j\boldsymbol{y}_j
			\right).
		\end{aligned}
		\label{eq:qf_parameters}
	\end{equation}
	
	\subsection{Variational E-Step: Precision-Multiplier Update}
	
	Using the moments of $q_j(\boldsymbol{f}_j)$, define
	\begin{equation}
		S_{ij}
		\triangleq
		(y_{ij}-\mu_{f,ij})^2
		+
		[\boldsymbol{C}_{f,j}]_{ii},
		\;
		R_{ij}
		\triangleq
		{S_{ij}}/{\sigma_j^2},
		\label{eq:residual_statistics}
	\end{equation}
	where $\mu_{f,ij}$ is the $i$th entry of
	$\boldsymbol{\mu}_{f,j}$. Let
	\begin{equation}
		\widetilde{a}
		\triangleq
		a+{1}/{2},
		\;
		\widetilde{b}_{ij}
		\triangleq
		b_j+{R_{ij}}/{2}.
		\label{eq:gamma_branch_updated_parameters}
	\end{equation}
	Applying \eqref{eq:variational_coordinate_update} to
	$\mathcal{I}_{ij}$ gives
	\begin{equation}
		\begin{aligned}
			q_j(\mathcal{I}_{ij})
			={}&
			(1-\Omega_{ij})
			\mathcal{G}
			\left(
			\mathcal{I}_{ij}
			\mid
			\widetilde{a},
			\widetilde{b}_{ij}
			\right)
			+
			\Omega_{ij}
			\delta(\mathcal{I}_{ij}-1),
		\end{aligned}
		\label{eq:qI_final_spike_gamma}
	\end{equation}
	where
	\begin{equation}
		\Omega_{ij}
		=
		\left[
		1+
		\frac{1-\theta_{ij}}{\theta_{ij}}
		\frac{\Gamma(\widetilde{a})}{\Gamma(a)}
		\frac{b_j^a}
		{\widetilde{b}_{ij}^{\widetilde{a}}}
		\exp\!\left(\frac{R_{ij}}{2}\right)
		\right]^{-1}.
		\label{eq:omega_closed_form}
	\end{equation}
	The corresponding robustness weight is
	\begin{equation}
		w_{ij}
		=
		\Omega_{ij}
		+
		(1-\Omega_{ij})
		{\widetilde{a}}/{\widetilde{b}_{ij}}.
		\label{eq:wij_final}
	\end{equation}
	
	\subsection{Generalized M-Step: Parameter Updates}
	
	With the variational quantities fixed,
	$\mathcal{Q}_j(\boldsymbol{\Theta}_j)$ separates into terms involving
	$\sigma_j^2$, $b_j$, and $(m_j,\boldsymbol{\kappa}_j)$.

	Collecting the terms that depend on $\sigma_j^2$ and setting their
	derivative to zero gives the MAP update
	\begin{equation}
		\sigma_j^2
		\leftarrow
		\frac{
			S_j+s_{0j}
		}{
			n+\nu_0+2
		}.
		\label{eq:sigma_point_estimate}
	\end{equation}
	where $S_j\triangleq\sum_{i=1}^{n}w_{ij}S_{ij}$.
	Similarly, the MAP update of the Gamma rate parameter is
	\begin{equation}
		b_j
		\leftarrow
		\frac{
			A-1+
			a\sum_{i=1}^{n}(1-\Omega_{ij})
		}{
			B+
			\sum_{i=1}^{n}
			(1-\Omega_{ij})
			\widetilde{a}/\widetilde{b}_{ij}
		}.
		\label{eq:bj_point_estimate}
	\end{equation}
	
	For fixed $\boldsymbol{\kappa}_j$, maximizing the GP-dependent terms
	of $\mathcal{Q}_j(\boldsymbol{\Theta}_j)$ with respect to $m_j$ gives
	\begin{equation}
		m_j
		\leftarrow
		{
			(\boldsymbol{1}_n^{\top}
			\boldsymbol{K}_j^{-1}
			\boldsymbol{\mu}_{f,j})
		}/{
			(\boldsymbol{1}_n^{\top}
			\boldsymbol{K}_j^{-1}
			\boldsymbol{1}_n)
		}.
		\label{eq:mj_point_estimate}
	\end{equation}
	
	For the kernel update, define
	\begin{equation}
		\begin{aligned}
			\boldsymbol{\phi}_j
			&\triangleq
			\begin{bmatrix}
				\log\sigma_{f,j} &
				\log\ell_{j1} &
				\cdots &
				\log\ell_{jd_x}
			\end{bmatrix}^{\top},
			\\
			\boldsymbol{\kappa}_j(\boldsymbol{\phi}_j)
			&\triangleq
			\begin{bmatrix}
				\exp(2\phi_{j1}) &
				\exp(\phi_{j2}) &
				\cdots &
				\exp(\phi_{j,d_x+1})
			\end{bmatrix}^{\top}.
		\end{aligned}
		\label{eq:log_kernel_parameterization}
	\end{equation}
	This logarithmic parameterization enforces positivity of the signal
	variance and length scales while allowing optimization over  $\boldsymbol{\phi}_j$.

	\begin{algorithm}[t]
		\caption{ASOR-GPR}
		\label{alg:independent_asor_gpr}
		
		\KwIn{
			Training data $(\boldsymbol{X},\boldsymbol{Y})$;
			prior hyperparameters $\{\mathcal{P}_j\}_{j=1}^{d_y}$;
			initial parameters $\{\boldsymbol{\Theta}_j\}_{j=1}^{d_y}$.
		}
		
		\KwOut{
			Converged independent-output ASOR-GPR models.
		}
		
		\For{$j\gets1$ \KwTo $d_y$}{
			
			Extract $\boldsymbol{y}_j$, initialize
			$\boldsymbol{\Theta}_j$, set $\boldsymbol{\phi}_j$ from
			$\boldsymbol{\kappa}_j$ using
			\eqref{eq:log_kernel_parameterization}, set
			$w_{ij}\gets1$, $i=1,\ldots,n$, and construct
			$\boldsymbol{K}_j$\;
			
			\Repeat{convergence}{
				
				\textbf{Variational E-step:}
				
				Update $q_j(\boldsymbol{f}_j)$ using
				\eqref{eq:asor_weight_and_precision} and
				\eqref{eq:qf_parameters}\;
				
				Update
				$\{q_j(\mathcal{I}_{ij}),\Omega_{ij},w_{ij}\}_{i=1}^{n}$
				using
				\eqref{eq:residual_statistics}--\eqref{eq:wij_final}\;
				
				\textbf{Generalized M-step:}
				
				Update $\sigma_j^2$, $b_j$, and $m_j$ using
				\eqref{eq:sigma_point_estimate}--\eqref{eq:mj_point_estimate}\;
				
				Update $\boldsymbol{\phi}_j$ by gradient descent with
				backtracking using
				\eqref{eq:kappa_negative_objective}--
				\eqref{eq:kernel_backtracking_rule}\;
				
				Update $\boldsymbol{\kappa}_j$ using
				\eqref{eq:log_kernel_parameterization} and reconstruct
				$\boldsymbol{K}_j$ using \eqref{eq:ard_se_kernel}\;
			}
			
			Perform a final variational E-step and store the
			converged quantities\;
		}
		
		\KwRet{The converged ASOR-GPR models}\;
	\end{algorithm}
	
	For fixed $m_j$, maximizing the terms of
	$\mathcal{Q}_j(\boldsymbol{\Theta}_j)$ that depend on
	$\boldsymbol{\kappa}_j$ through $\boldsymbol{K}_j$ is equivalent to
	minimizing
	\begin{equation}
		\mathcal{J}_j(\boldsymbol{\phi}_j)
		\triangleq
		\frac{1}{2}
		\operatorname{tr}
		\left(
		\boldsymbol{K}_j^{-1}
		\boldsymbol{D}_j
		\right)
		+
		\frac{1}{2}
		\log
		\left|
		\boldsymbol{K}_j
		\right|.
		\label{eq:kappa_negative_objective}
	\end{equation}
	where
	\begin{equation}
		\boldsymbol{D}_j
		\triangleq
		\boldsymbol{C}_{f,j}
		+
		\left(
		\boldsymbol{\mu}_{f,j}
		-
		m_j\boldsymbol{1}_n
		\right)
		\left(
		\boldsymbol{\mu}_{f,j}
		-
		m_j\boldsymbol{1}_n
		\right)^{\top}.
		\label{eq:dj_definition}
	\end{equation}
	To obtain a descent direction for minimizing
	$\mathcal{J}_j(\boldsymbol{\phi}_j)$, its gradient components are
	\begin{equation}
		\frac{\partial\mathcal{J}_j}
		{\partial\phi_{jr}}
		=
		\frac{1}{2}
		\operatorname{tr}
		\left[
		\left(
		\boldsymbol{K}_j^{-1}
		-
		\boldsymbol{K}_j^{-1}
		\boldsymbol{D}_j
		\boldsymbol{K}_j^{-1}
		\right)
		\frac{\partial\boldsymbol{K}_j}
		{\partial\phi_{jr}}
		\right].
		\label{eq:log_kernel_gradient}
	\end{equation}
	The required kernel derivatives are
	\begin{equation}
		\begin{aligned}
			\frac{\partial\boldsymbol{K}_j}
			{\partial\phi_{j1}}
			=
			2\boldsymbol{K}_j,\quad
			\left[
			\frac{\partial\boldsymbol{K}_j}
			{\partial\phi_{j,r+1}}
			\right]_{i\ell}
			=
			[\boldsymbol{K}_j]_{i\ell}
			\frac{(x_{ir}-x_{\ell r})^2}{\ell_{jr}^2}.
		\end{aligned}
		\label{eq:kernel_log_derivatives}
	\end{equation}

	% For a trial step size $\alpha_j>0$, the normalized-gradient candidate is
	% \begin{equation}
	% 	\boldsymbol{\phi}_j^{\mathrm{cand}}
	% 	=
	% 	\boldsymbol{\phi}_j^{\mathrm{old}}
	% 	-
	% 	\alpha_j
	% 	\frac{\boldsymbol{g}_j}
	% 	{\max\{1,\|\boldsymbol{g}_j\|_2\}},
	% 	\label{eq:normalized_kernel_gradient_step}
	% \end{equation}
	% where
	% $\boldsymbol{g}_j\triangleq
	% \nabla_{\boldsymbol{\phi}_j}\mathcal{J}_j$. A backtracking line search reduces $\alpha_j$ until

%     \begin{equation}
% 	\mathcal{J}_j(\boldsymbol{\phi}_j^{\mathrm{cand}})
% 	\leq
% 	\mathcal{J}_j(\boldsymbol{\phi}_j^{\mathrm{old}})
% 	+
% 	c_1\alpha_j
% 	\boldsymbol{g}_j^{\top}\boldsymbol{d}_j,
% 	\label{eq:kernel_backtracking_rule}
% \end{equation}

For a trial step size $\alpha_j>0$, the normalized-gradient candidate is
\begin{equation}
	\boldsymbol{\phi}_j^{\mathrm{cand}}
	=
	\boldsymbol{\phi}_j^{\mathrm{old}}
	-
	\alpha_j
	\frac{\boldsymbol{g}_j}
	{\max\{1,\|\boldsymbol{g}_j\|_2\}},
	\label{eq:normalized_kernel_gradient_step}
\end{equation}
where
$\boldsymbol{g}_j\triangleq
\nabla_{\boldsymbol{\phi}_j}\mathcal{J}_j$.
A backtracking line search halves $\alpha_j$ until the Armijo condition
\begin{equation}
	\mathcal{J}_j(\boldsymbol{\phi}_j^{\mathrm{cand}})
	\leq
	\mathcal{J}_j(\boldsymbol{\phi}_j^{\mathrm{old}})
	-
	c_1\alpha_j
	\frac{\|\boldsymbol{g}_j\|_2^2}
	{\max\{1,\|\boldsymbol{g}_j\|_2\}}
	\label{eq:kernel_backtracking_rule}
\end{equation}
is satisfied, where $c_1$ is the Armijo constant. The accepted value is
	$\boldsymbol{\phi}_j\leftarrow
	\boldsymbol{\phi}_j^{\mathrm{cand}}$,
	followed by
	$\boldsymbol{\kappa}_j\leftarrow
	\boldsymbol{\kappa}_j(\boldsymbol{\phi}_j)$ and
	$\boldsymbol{K}_j\leftarrow
	\boldsymbol{K}_j(\boldsymbol{\kappa}_j)$.

Iterations terminate when the relative change in the weighted residual
diagnostic $S_j$ falls below a prescribed
tolerance or the maximum iteration count is reached.
	
	\subsection{Posterior Prediction}
	
	Let
	$\widehat{m}_j$, $\widehat{\boldsymbol{\kappa}}_j$, and
	$\widehat{\sigma}_j^2$ denote the learned parameters, and let
	\begin{equation}
		q_j(\boldsymbol{f}_j)
		=
		\mathcal{N}
		\left(
		\boldsymbol{f}_j
		\mid
		\widehat{\boldsymbol{\mu}}_{f,j},
		\widehat{\boldsymbol{C}}_{f,j}
		\right).
		\label{eq:final_latent_posterior}
	\end{equation}
	
	For a test input $\boldsymbol{x}^{*}$, let
	$f_{*j}\triangleq f_j(\boldsymbol{x}^{*})$ and define
	\begin{equation}
		\begin{aligned}
			\widehat{\boldsymbol{K}}_j
			&\triangleq
			\boldsymbol{K}_j
			\left(
			\widehat{\boldsymbol{\kappa}}_j
			\right),
			\qquad
			\widehat{\boldsymbol{k}}_{*j}
			\triangleq
			\left[
			k_j
			\left(
			\boldsymbol{x}^{i},
			\boldsymbol{x}^{*};
			\widehat{\boldsymbol{\kappa}}_j
			\right)
			\right]_{i=1}^{n},
			\\
			\widehat{k}_{**j}
			&\triangleq
			k_j
			\left(
			\boldsymbol{x}^{*},
			\boldsymbol{x}^{*};
			\widehat{\boldsymbol{\kappa}}_j
			\right).
		\end{aligned}
	\end{equation}
	Marginalizing the conditional GP distribution over
	\eqref{eq:final_latent_posterior} gives
	\begin{equation}
		\begin{aligned}
			q_j(f_{*j})
			&=
			\mathcal{N}
			\left(
			f_{*j}
			\mid
			\mu_{*j},
			v_{*j}
			\right),
			\\
			\mu_{*j}
			&=
			\widehat{m}_j
			+
			\widehat{\boldsymbol{k}}_{*j}^{\top}
			\widehat{\boldsymbol{K}}_j^{-1}
			\left(
			\widehat{\boldsymbol{\mu}}_{f,j}
			-
			\widehat{m}_j\boldsymbol{1}_n
			\right),
			\\
			v_{*j}
			&=
			\widehat{k}_{**j}
			-
			\widehat{\boldsymbol{k}}_{*j}^{\top}
			\widehat{\boldsymbol{K}}_j^{-1}
			\widehat{\boldsymbol{k}}_{*j}
			+
			\widehat{\boldsymbol{k}}_{*j}^{\top}
			\widehat{\boldsymbol{K}}_j^{-1}
			\widehat{\boldsymbol{C}}_{f,j}
			\widehat{\boldsymbol{K}}_j^{-1}
			\widehat{\boldsymbol{k}}_{*j}.
		\end{aligned}
		\label{eq:predictive_latent_distribution}
	\end{equation}
	
	For a nominal future observation $y_{*j}$, the predictive distribution
	is obtained as
	\begin{equation}
		q_j(y_{*j})
		=
		\mathcal{N}
		\left(
		y_{*j}
		\mid
		\mu_{*j},
		v_{*j}+\widehat{\sigma}_j^2
		\right).
		\label{eq:noisy_predictive_distribution}
	\end{equation}
	
	\section{Numerical Experiments}
\label{sec:numerical_experiments}
     \begin{figure*}[!t]
		\centering
		\subfloat[Synthetic data: uniform outliers]{
			\includegraphics[width=0.3\textwidth]{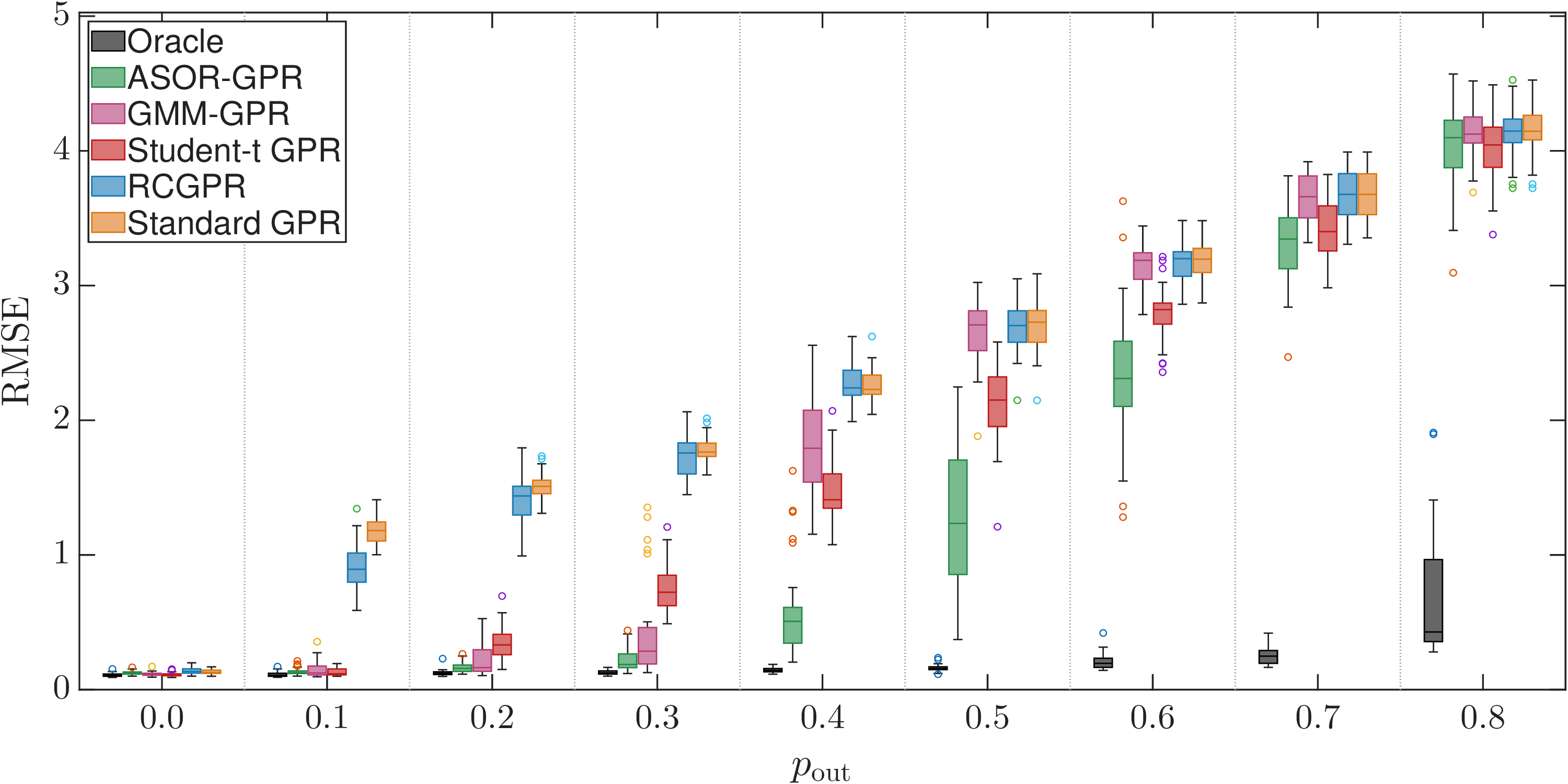}
			\label{fig:synthetic_uniform}
		}
		\hfill
		\subfloat[Air Quality: uniform outliers]{
			\includegraphics[width=0.3\textwidth]{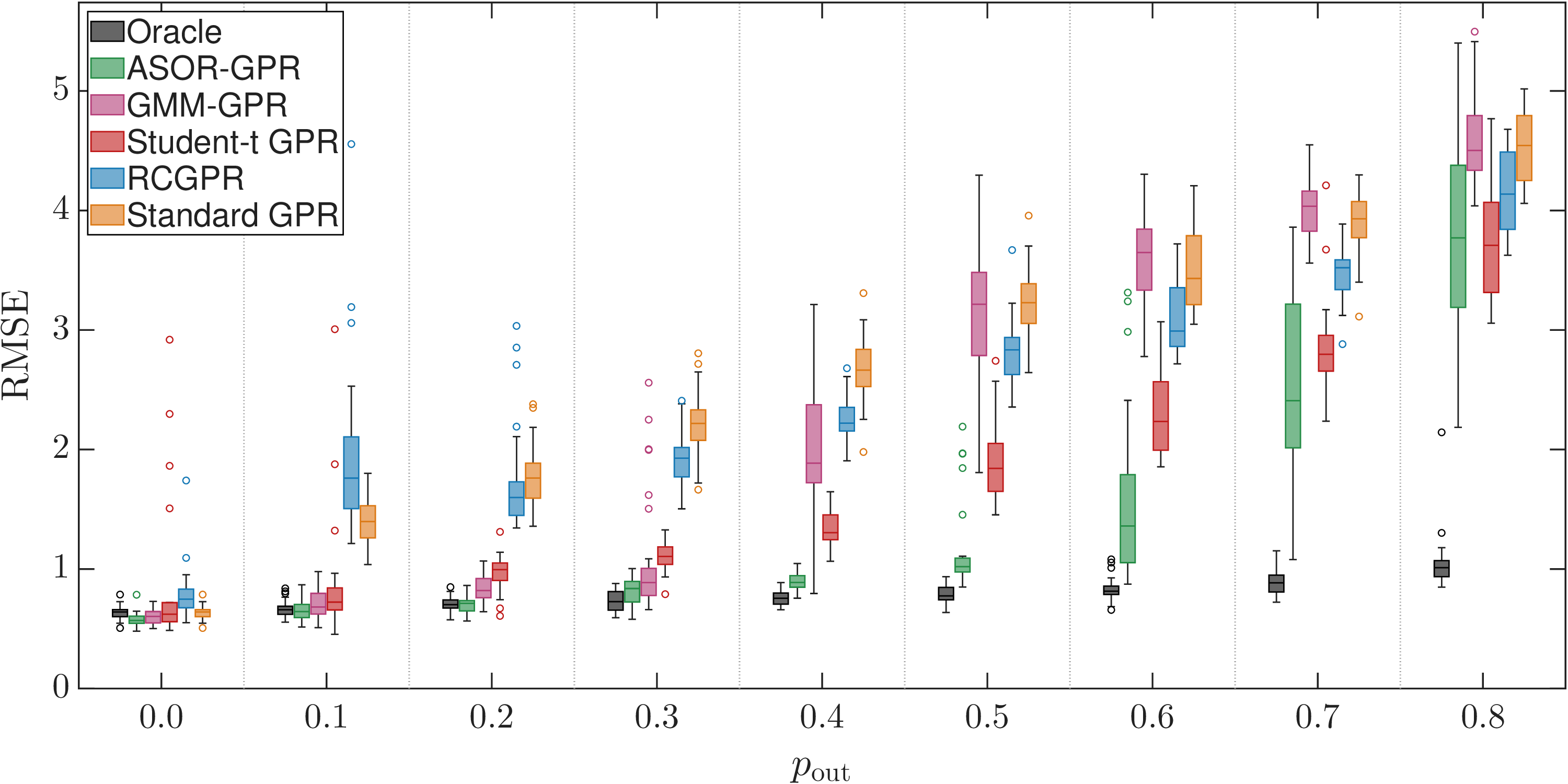}
			\label{fig:airquality_uniform}
		}
		\hfill
		\subfloat[Energy Efficiency: Gaussian outliers]{
			\includegraphics[width=0.3\textwidth]{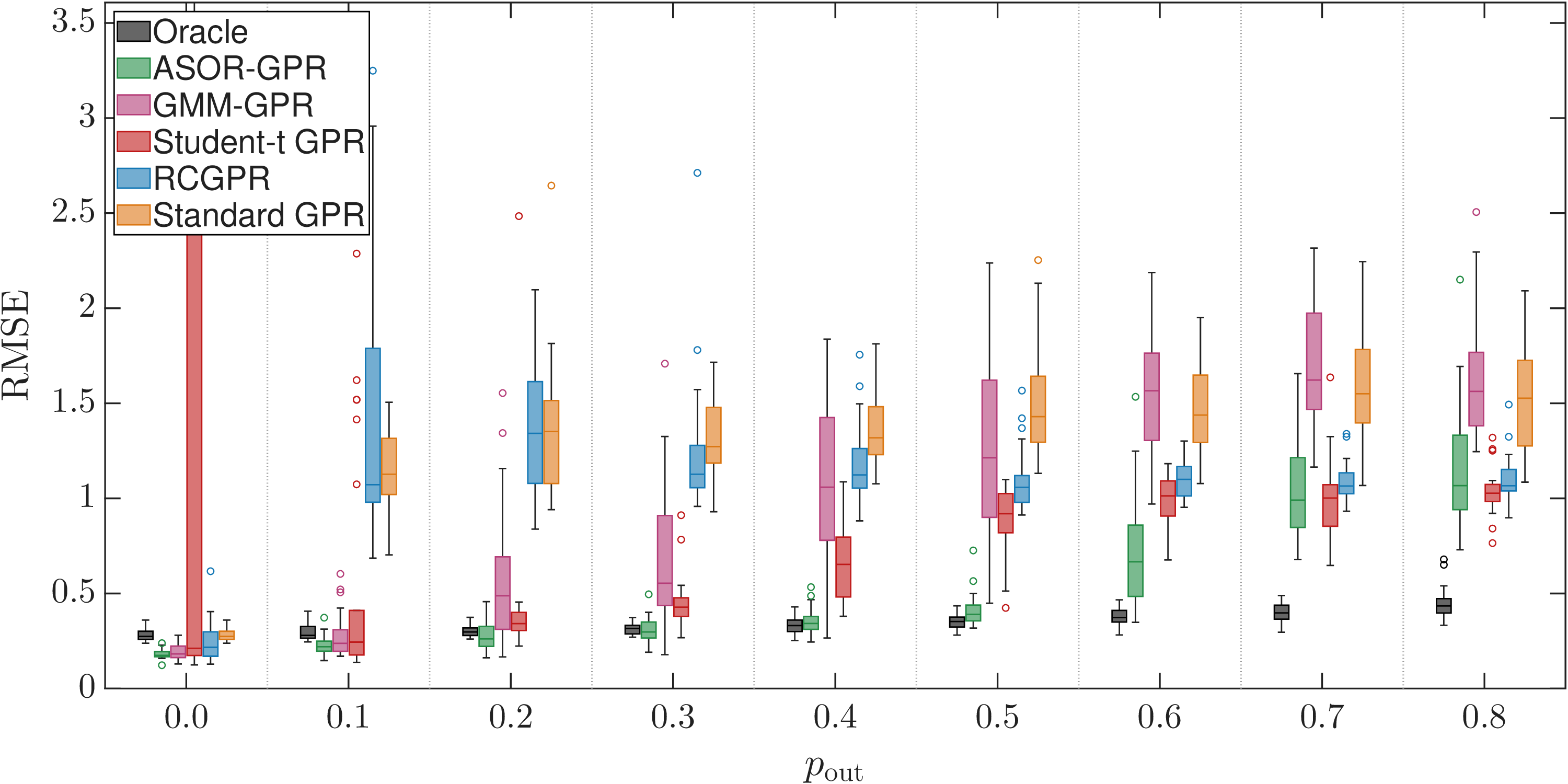}
			\label{fig:energy_gaussian}
		}
		\caption{Prediction performance under different outlier settings:
			(a) synthetic data with uniform outliers over $[0,10]$;
			(b) Air Quality data with uniform outliers over $[-5,15]$; and
			(c) Energy Efficiency data with zero-mean Gaussian outliers of
			variance $25$.}
		\label{fig:prediction_results}
		\vspace{0mm}
	\end{figure*}

All experiments were conducted in MATLAB R2026a on an AMD Ryzen 9
9950X desktop with 64 GB RAM. We compare ASOR-GPR with standard GPR
\cite{rasmussen2006gaussian}, robust conjugate GPR (RCGPR)
\cite{altamirano2024robust}, Student-$t$ GPR
\cite{jylanki2011robust}, Gaussian-mixture-likelihood GPR (GMM-GPR)
\cite{daemi2019gaussian}, and an Oracle GPR supplied with all available ground-truth
contamination information.

All methods use independent scalar-output models with the ARD
squared-exponential kernel, initialized with
$\sigma_{f,j}^{(0)}=1$ and
$\ell_{jr}^{(0)}=0.2d_{\mathrm{med}}$, where $d_{\mathrm{med}}$ is
the median pairwise training-input distance. All non-oracle methods are initialized with nominal-noise variance $0.5$.
For synthetic data, this is deliberately misspecified relative to the true
variance $0.25$. For the real datasets, the true nominal-noise
variance is unknown. Predictive experiments use
	$n_{\mathrm{train}}=n_{\mathrm{test}}=100$.

Outliers are injected entry-wise only into the training targets over
30 Monte Carlo trials. Root mean square error (RMSE) for synthetic data is
evaluated against the known latent function at the test inputs. For
real data, the original observations are treated as nominal, with
outliers injected only into the training targets and RMSE evaluated
against the unmodified test targets, which retain their inherent
observation noise.

Baseline-specific parameters follow their respective references unless
otherwise stated. ASOR-GPR uses $a=1$, $\theta_{ij}=0.5$, $A=10$,
$B=0.45$, $\nu_0=3$, $b_j^{(0)}=20$, and $s_{0j}=1.25$ in all real-data and synthetic experiments. Up to 1000 variational iterations are run,
terminating when relative change in $S_j$ falls below  $10^{-5}$. Kernel hyperparameters are optimized by normalized gradient descent with
Armijo backtracking using initial step $\alpha_j=0.05$, Armijo constant
$c_1=10^{-4}$, minimum step $\alpha_{\min}=10^{-7}$, and at most 25
inner iterations.

For the synthetic experiment, $d_x=2$ and $d_y=4$. Inputs are
Latin-hypercube sampled uniformly over $[-3,3]^2$. Defining
$z_{ir}=(x_{ir}+3)/6$, $r=1,2$, the clean outputs are generated as
\begin{equation}
	\begin{aligned}
		g_j(\boldsymbol{x}^{i})
		={}&
		10\sin(\pi z_{i1}z_{i2})+5
		+0.75\sin\!\left((0.8+0.1j)x_{i1}\right)
		\\[-1mm]
		&+
		0.50\cos\!\left((0.6+0.05j)x_{i2}\right)
		+0.10j\,x_{i1},
		\\
		\widetilde{\boldsymbol{g}}(\boldsymbol{x}^{i})
		={}&
		\boldsymbol{C}_y^{\top}
		\begin{bmatrix}
			g_1(\boldsymbol{x}^{i}) & \cdots &
			g_{d_y}(\boldsymbol{x}^{i})
		\end{bmatrix}^{\top},
		\\
		\boldsymbol{C}_y
		={}&
		\boldsymbol{I}_{d_y}
		+0.1
		\left(
		\boldsymbol{1}_{d_y}\boldsymbol{1}_{d_y}^{\top}
		-\boldsymbol{I}_{d_y}
		\right).
	\end{aligned}
	\label{eq:synthetic_generation}
\end{equation}
Here, $\widetilde{\boldsymbol{g}}(\boldsymbol{x}^{i})$ denotes the
mixed clean output vector. The outputs are standardized to zero mean and unit variance before adding nominal Gaussian noise and entry-wise contamination, with
$p_{\mathrm{out}}\in\{0,0.1,\ldots,0.8\}$.

Fig.~\ref{fig:synthetic_uniform} considers outliers drawn uniformly
from $[0,10]$. ASOR-GPR remains competitive across the contamination
range and achieves the lowest RMSE among the compared
non-oracle methods at most contamination levels. For real-data evaluation, we use the Air Quality
	\cite{air_quality_360} and Energy Efficiency
	\cite{energy_efficiency_242} datasets, standardized using training-set
	statistics. 
	Using the same probability sweep,
	Fig.~\ref{fig:airquality_uniform} considers uniform outliers over
	$[-5,15]$, while Fig.~\ref{fig:energy_gaussian} considers zero-mean
	Gaussian outliers with variance $25$. ASOR-GPR provides clear gains under asymmetric contamination in
	Fig.~\ref{fig:airquality_uniform} and remains competitive with robust
	likelihood-based methods under symmetric Gaussian outliers in
	Fig.~\ref{fig:energy_gaussian}. Similar trends are observed over a
	wider range of contamination intensity. This reflects the effectiveness of the proposed adaptive inference of
observation-specific precisions and nominal/outlier associations.

% \begin{table}[!h]
% 	\centering
% 	\caption{Median model-fitting time (s).}
% 	\label{tab:runtime}
% 	\scriptsize
% 	\setlength{\tabcolsep}{5.2pt}
% 	\renewcommand{\arraystretch}{0.86}
% 	\begin{tabular}{lcccc}
% 		\hline
% 		\textbf{Method} &
% 		\textbf{100} &
% 		\textbf{200} &
% 		\textbf{300} &
% 		\textbf{400} \\
% 		\hline
% 		Oracle
% 		& 0.018594 & 0.063985 & 0.15355 & 0.57685 \\
		
% 		ASOR-GPR
% 		& 5.5290 & 14.087 & 19.213 & 27.127 \\
		
% 		GMM-GPR
% 		& 7.3416 & 20.304 & 44.775 & 100.17 \\
		
% 		Student-$t$ GPR
% 		& 1.6548 & 2.6900 & 5.8321 & 14.558 \\
		
% 		RCGPR
% 		& 0.067125 & 0.30185 & 0.74591 & 3.2568 \\
		
% 		Std. GPR
% 		& 0.031718 & 0.11725 & 0.26715 & 1.1077 \\
% 		\hline
% 	\end{tabular}
% 	\vspace{-2mm}
% \end{table}

\begin{table}[!h]
	\centering
	\caption{Median serial model-fitting time (s) over 30 Monte Carlo trials.}
	\label{tab:runtime}
	\scriptsize
	\setlength{\tabcolsep}{3.0pt}
	\renewcommand{\arraystretch}{0.90}
	\begin{tabular}{lccccc}
		\hline
		\textbf{Method} &
		\multicolumn{5}{c}{$\boldsymbol{n_{\mathrm{train}}}$} \\
		\cline{2-6}
		&
		\textbf{200} &
		\textbf{400} &
		\textbf{600} &
		\textbf{800} &
		\textbf{1000} \\
		\hline
		Oracle GPR
		& 0.10 & 0.70 & 1.99 & 4.37 & 6.76 \\

		Std. GPR
		& 0.19 & 1.46 & 4.24 & 9.43 & 15.05 \\

		RCGPR
		& 0.50 & 4.74 & 15.58 & 36.45 & 62.73 \\

		Student-\(t\) GPR
		& 3.25 & 16.23 & 41.39 & 80.75 & 123.53 \\

		ASOR-GPR
		& 24.60 & 31.13 & 85.52 & 212.41 & 335.96 \\

		GMM-GPR
		& 31.96 & 151.23 & 376.66 & 644.76 & 963.38 \\
		\hline
	\end{tabular}
	\vspace{-2mm}
\end{table}
For runtime evaluation, Table~\ref{tab:runtime} reports median
model-fitting times with increasing training-set size at
$p_{\mathrm{out}}=0.4$ under zero-mean Gaussian contamination with
variance $10$. Oracle GPR and standard GPR have the lowest runtimes, with the Oracle
benefiting from known contamination information. RCGPR and Student-$t$
GPR incur moderate computational cost. ASOR-GPR and GMM-GPR exhibit
the highest runtimes, with ASOR-GPR consistently faster than GMM-GPR
over the reported training sizes. This relative computational behavior
remains broadly consistent across different contamination
probabilities. All methods involve dense covariance-matrix operations
and retain cubic complexity, while runtime differences
arise from their method-specific inference procedures.
    
	\section{Conclusion}
	We proposed an outlier-robust GPR approach that embeds
	adaptive observation-specific precision modeling within a hierarchical generative framework. Variational generalized-EM inference enables
observation-specific contamination effects and the underlying GP model
to be learned jointly. Results on synthetic and real datasets show that ASOR-GPR achieves
competitive predictive performance with enhanced robustness across a
range of contamination settings, while preserving the cubic computational scaling of the standard GPR.

	\bibliographystyle{IEEEtran}
	\bibliography{references}
	
\end{document}